\documentclass{article}

\usepackage{caption}
\usepackage{subcaption}
\usepackage{multirow}

\usepackage{hyperref}

\usepackage[accepted]{icml2024}

\usepackage{amsmath, bm, bbm}
\usepackage{amssymb, amsthm}
\usepackage{mathtools}

\usepackage[capitalize,noabbrev]{cleveref}

\theoremstyle{plain}

\theoremstyle{definition}

\theoremstyle{remark}

\icmltitlerunning{Quo Vadis, Time Series Anomaly Detection?}

\begin{document}
\twocolumn[
\icmltitle{Position: Quo Vadis, Unsupervised Time Series Anomaly Detection?}




\begin{icmlauthorlist}
\icmlauthor{M. Saquib Sarfraz}{comp,sch}
\icmlauthor{Mei-Yen Chen}{comp}
\icmlauthor{Lukas Layer}{comp}
\icmlauthor{Kunyu Peng}{sch}
\icmlauthor{Marios Koulakis}{sch}
\end{icmlauthorlist}

\icmlaffiliation{sch}{Karlsruhe Institute of Technology, Karlsruhe, Germany}
\icmlaffiliation{comp}{Mercedes-Benz Tech Innovation, Ulm, Germany}

\icmlcorrespondingauthor{M. Saquib Sarfraz}{saquibsarfraz@gmail.com}

\icmlkeywords{Machine Learning, time series, anomaly detection, multivariate time series}

\vskip 0.3in
]



\printAffiliationsAndNotice{}  

\begin{abstract}
The current state of machine learning scholarship in Timeseries Anomaly Detection (TAD) is plagued by the persistent use of flawed evaluation metrics, inconsistent benchmarking practices, and a lack of proper justification for the choices made in novel deep learning-based model designs. Our paper presents a critical analysis of the status quo in TAD, revealing the misleading track of current research and highlighting problematic methods, and evaluation practices. \textbf{Our position advocates for a shift in focus from solely pursuing novel model designs to improving benchmarking practices, creating non-trivial datasets, and critically evaluating the utility of complex methods against simpler baselines}. Our findings demonstrate the need for rigorous evaluation protocols, the creation of simple baselines, and the revelation that state-of-the-art deep anomaly detection models effectively learn linear mappings. These findings suggest the need for more exploration and development of simple and interpretable TAD methods. The increment of model complexity in the state-of-the-art deep-learning based models unfortunately offers very little improvement. We offer insights and suggestions for the field to move forward.
\end{abstract}

\section{Introduction}
\label{sec:intro}

Time series anomaly detection (TAD) is an active field of machine learning with applications across multiple industries.
For instance, many real-world systems such as vehicles, manufacturing plants, robots, and  patient monitoring systems, involve a large number of interconnected sensors producing a great amount of data over time that can be used to detect anomalous behaviour. 
The anomalies can manifest as single irregular points or groups of such points whose interpretation as anomalous might depend on the system's operational history or on the inter-connectivity among sub-modules.

Given the complexity of the problem and inspired from the successes in other areas, such as natural language or audio processing, many state-of-the-art deep-learning architectures have been adjusted and applied to it. 
Such approaches aim to learn a latent representation of the normal time-series data, e.g. LSTM \cite{lstmvae}, Transformer \cite{tuli2022tranad, xu2022anomaly}, and sometimes explicitly model the inter-dependency among the sub-components in the system, e.g. graph neural networks \cite{gdn, zekaietal-gta-2021}.
Based on the assumption that the anomalies constitute unseen patterns which will not be modelled during reconstruction of the series from the model, the difference between the original and reconstructed series is used to detect them.

Although it is well intended, this line of research has never provided evidence of the necessity of deep-learning, which has been challenged namely in \citeauthor{Audibert_2022_Question_DL} (\citeyear{Audibert_2022_Question_DL}). The state-of-the-art (SOTA) deep-learning approaches proceeded to introduce models of increased complexity using questionable validation processes.
Those processes involve unsuitable benchmark datasets \cite{Wu_Keogh_2022} and, most harmful to this field, the use of flawed evaluation protocols \cite{kim2022TADevaluation}.
The protocol which introduced the most pitfalls is the point adjustment (PA) applied on the point-wise F1 score which practically favors noisy predictions.
It was gradually introduced in a series of papers~\cite{unsupvariationae, usad, NEURIPS2020_97e401a0, smd_dataset} with the original intention of calibrating the anomaly detection threshold on a hold-out dataset, but it was subsequently demonstrated in \citeauthor{kim2022TADevaluation} (\citeyear{kim2022TADevaluation}) that uniformly random predictions outperform SOTA methods and their performance tends to one as the average length of the anomalies increases.
Although using the standard F1 score without point-adjust avoids those pitfalls, it still leaves a gap by only focusing on point-wise time-stamp level detection versus anomaly instance level detection, which led to the introduction of new complementary range-based metrics such as the ones in \citeauthor{tatbul2018precision} (\citeyear{tatbul2018precision}), \citeauthor{wagner2023timesead} (\citeyear{wagner2023timesead}).

The goal of this paper is to guide the TAD community towards more meaningful progress through rigorous benchmarking practices and a focus on studying the utility of their models by drawing useful but simple baselines. We achieve this with the following contributions:
1.) We introduce simple and effective baselines and demonstrate that they perform on par or better than the SOTA methods, thus challenging the efficiency and effectiveness of increasing model complexity to solve TAD problems.
2.) We reinforce this position by reducing trained SOTA models to linear models which are distillations of them but still perform on par. Thus from the point of view of the TAD task on the current datasets, those models perform roughly a linear separation of the anomalies from the nominal data.

Our code\footnote{\scriptsize Code: \url{https://github.com/ssarfraz/QuoVadisTAD}} is available on GitHub to easily run the baselines and benchmarks.

\section{Related Work}
\label{sec:related}

Anomaly detection in time series data has been extensively studied, with methods ranging from univariate to multivariate and including complex deep-learning models~\cite{madgan, mscred, Zhao2020MultivariateTA, OmniAnomaly, dagmm, lstmndt, gdn, zekaietal-gta-2021}. 
These models are trained to forecast or reconstruct presumed normal system states and then deployed to detect anomalies in unseen test datasets. 
The anomaly score defined as the magnitude of prediction or reconstruction errors serves as an indicator of abnormality at each time stamp. 
Model performance is often evaluated as a binary classification problem, with the anomaly scores thresholded into binary labels. 
A comprehensive review of anomaly detection methods can be found in~\cite{SchmidlEtAl2022Anomaly, tad_review_2020}. 

\textbf{Classical machine learning methods:} A basic approach to anomaly detection in time-series data involves treating sample points of each sensor as independent and using classical statistical methods on the individual univariate series.
For instance, regression models are used for the prediction from other sensor measurements \cite{Salem_svm_regression}. 
Principal Component Analysis (PCA) is utilized for dimensionality reduction and reconstruction \cite{Shyu2006_PCA}. 
Other methods for anomaly detection on time series data take temporal dependency or correlation among sensors into account. 
These include modeling families of hidden Markov chains \cite{Patcha_hidden_markov} or graph theory \cite{GraphAn_2020}. 
Signal transformation \cite{Kanarachos_wavelets}, isolation forest \cite{Bandaragoda_isoforest, Liu_ieee_isoforest}, Auto-Regressive Integrated Moving Average (ARIMA) \cite{Yaacob_arima} and clustering \cite{Angiulli_KNN, sand_2021, Tran_rdbo}. 
Time-series discord discovery has recently emerged as a favored choice for univariate data analysis. 
A recent method MERLIN~\cite{Nakamura2020MERLINPD} is considered to be the state-of-the-art for univariate anomaly detection, as it iteratively varies the length of a subsequence and searches for those that are greatly different from their nearest neighbors as candidates of abnormality. Also see \cite{paparrizos2022tsb} for a comprehensive performance comparison of different classical TAD methods on univariate data.

\textbf{Deep learning methods:} 
Anomaly in time series might be hidden in peculiar dependencies among sub-modules in a system or over its operation history that are hard to detect with manual feature engineering. 
Modern deep-learning models that can learn temporal dependency via recursive networks (e.g. LSTM) or attention mechanisms (e.g. Transformer) or by explicitly representing the correlation among sensors (e.g. Graph Neural Networks) have been proposed as the cutting-edge methods for TAD. 
For instance, LSTM-VAE \cite{lstmvae} used a variational autoencoder that is based on LSTM and reconstructs the test data with variational inferences. 
DAGMM \cite{dagmm} utilized deep autoencoders and Gaussian mixture model to jointly model a low-dimensional representation which is then used to reconstruct each time stamp. 
It computes the reconstruction error for anomaly detection. 
OmniAnomaly\cite{OmniAnomaly} modeled the time series data as stochastic random process with variational autoencoders (VAE) and established reconstruction likelihood as an anomaly score. 
Another approach, USAD \cite{usad}, introduced a two-phase training paradigm in which two autoencoders and two decoders are trained under the adversarial game-style.
Among the more recent methods that currently represent the state-of-the-art deep models on anomaly detection are GDN \cite{gdn} and TranAD~\cite{tuli2022tranad}.
GDN \cite{gdn} models the inter-connectivity among sensors as a graph and used graph attention network to forecast the sensor measurement. 
The deviation between true observation and model predictions is then used to quantify anomalies.
TranAD \cite{tuli2022tranad} is a transformer based approach that proposed a new transformer architecture for anomaly detection. 
It introduced several components with a two transformer-based encoder and decoders using multi-head attention blocks. 
The approach then proposed a two-phase training scheme utilizing adversarial and meta learning procedures. 
Another recent transformer based approach Anomaly Transformer \cite{xu2022anomaly} introduced a new attention block and a min-max loss which helps learn two separate series associations, one prior which aims to capture local associations which in cases of anomaly would be caused by the continuity around it and series associations which should encode deeper information about the temporal context. Overall both methods results in complicated schemes.
A similar approach in designing a transformer based model along with meta learning objectives and optimal transport has been presented in \cite{puad_icml}.

Aside from the anomaly detection approaches, many efforts has been put in creating useful anomaly detection benchmarks. 
Some recent studies, for instance~\cite{Wu_Keogh_2022} have shown how some of these datasets suffer from potential flaws, such as triviality, unrealistic density of anomaly, or mislabeling. 
\section{Methods}
\label{sec:method}
Among the numerous anomaly detection approaches presented in the past, there is often something consistent - they tend to overlook simpler baselines in pursuit of novelty.
This leads to overly complex engineered solutions without much utility and a good rationale.
Towards this end, we propose simple methods that exceed the performance of current best-published anomaly detection approaches. 
As a result, these baselines help us to understand the complexity of the underlying problem and provide a solid foundation for further investigation.
Of note, our contribution is properly setting up these known methods and creating a set of strong baselines.
\subsection{Preliminaries} 
We introduce some notations which are used to formally define the task of unsupervised TAD and describe the methods used. 
The training data consist of a time series  $\mathbf{X}=[\mathbf{x}_1,\ldots \mathbf{x}_T]\in\mathbb{R}^{T\cdot F}$ which only contains non-anomalous timestamps. 
Here $T$ is the number of timestamps and $F$ the number of features. 
The test set, $\hat{\mathbf{X}}=[\hat{\mathbf{x}}_1, \ldots \hat{\mathbf{x}}_{\hat{T}}]\in\mathbb{R}^{\hat{T}\cdot F}$ contains both normal and anomalous timestamps and $\hat{\mathbf{y}}=[\hat{y}_1, \ldots, \hat{y}_{\hat{T}}]\in\{0, 1\}^{\hat{T}}$ represents their labels, where $\hat{y}_t = 0$ denotes a normal and $\hat{y}_t = 1$ an anomalous timestamp $t$. 
Then the task of anomaly detection is to select a function $f_{\bm{\theta}}: \mathbf{X} \rightarrow \mathbb{R}$ such that $f_{\bm{\theta}}(\mathbf{x}_t)=\tilde{y}_t$ estimates the anomaly value $\hat{y}_t$\footnote{\scriptsize The range of $\tilde{y}_t$ values may differ from $\hat{y}_t\in {0, 1}$, necessitating thresholding before obtaining actual predictions. 
Typically, the threshold which yields the best score on the training or validation data is selected.}. 
The (potentially empty) set of parameters $\bm{\theta}$ is estimated using the training data $\mathbf{X}$. 
In most methods, usually an intermediate error vector function $err_{\bm{\theta}}: \mathbf{X} \rightarrow \mathbb{R}^F$ is estimated which computes vectors representing an error along all sensors, we also denote by $\mathbf{E} = err_{\bm{\theta}}(\hat{\mathbf{X}})$ the predicted test error vectors. 

The error vectors $\mathbf{E}$ estimated from any of the methods provide a measure of the deviation of the test features from normality. 
Normalization of error vectors sometimes is necessary before detecting anomalies due to variations in error behavior across sensors. Two normalization methods are often used: scaling using robust statistics such as median and inter-quartile range~\cite{gdn} and scaling using mean and standard deviation. The choice of normalization approach can impact anomaly detection accuracy, and careful consideration should be given to the selected method. 
The impact of error vector normalization on datasets is demonstrated through an ablation study in section~\ref{subsec:abl-norm}.
Once the error vectors are normalized, the final output is a measure of the vector sizes. Given that we are working on the anomaly detection scenario, the most fitting metric is $L^{\infty}$ which computes the largest absolute error between the different sensors, $\|\mathbf{e}_t\|_{\infty}=\max\limits_{i\leq F}\{|e_t^i|\}$.

\subsection{Proposed simple and effective baselines}
\textbf{Sensor range deviation:} 
The range of sensor values observed during normal operation can be useful in identifying out-of-distribution (OOD) samples. 
Anomalies in time series data can occur when the sensor values deviate from their usual range. 
Therefore, if the sensor values in a test data point fall outside the observed range, it may indicate the presence of an anomaly.
Formally this is defined as: 
$$
f(\hat{\mathbf{x}}_t) = \left\{
\begin{array}{ll}
      0 & if\ \hat{\mathbf{x}}_t\in [\min(\mathbf{X}), \max(\mathbf{X})] \\
      1 & otherwise \\
\end{array} 
\right\}
$$

This represents a minimum level of detection performance that any advanced method should be able to surpass. 

\textbf{L2-norm: Magnitude of the observed time stamp:}
In the case of multivariate time series data, the magnitude of the vector at a particular timestamp may serve as a relevant statistic for detecting OOD samples. 
This can be easily computed by taking the L2-norm of the vector, thus $f(\hat{\mathbf{x}}_t) = \|\hat{\mathbf{x}}_t\|_2$. By using the magnitude as an anomaly score, we have discovered that it can be an effective and robust baseline for identifying anomalies in multivariate datasets.

\textbf{NN-distance: Nearest neighbor distance to the normal training data:}
A sample that deviates from normal data should have a greater distance from it. 
Therefore, using the nearest-neighbor distance between each test time-stamp and the train data as an anomaly score can serve as a reliable baseline. 
In fact, in many cases, this method outperforms several state-of-the-art techniques.

\begin{figure*}[t]
\centerline{\includegraphics[width=0.73\textwidth]{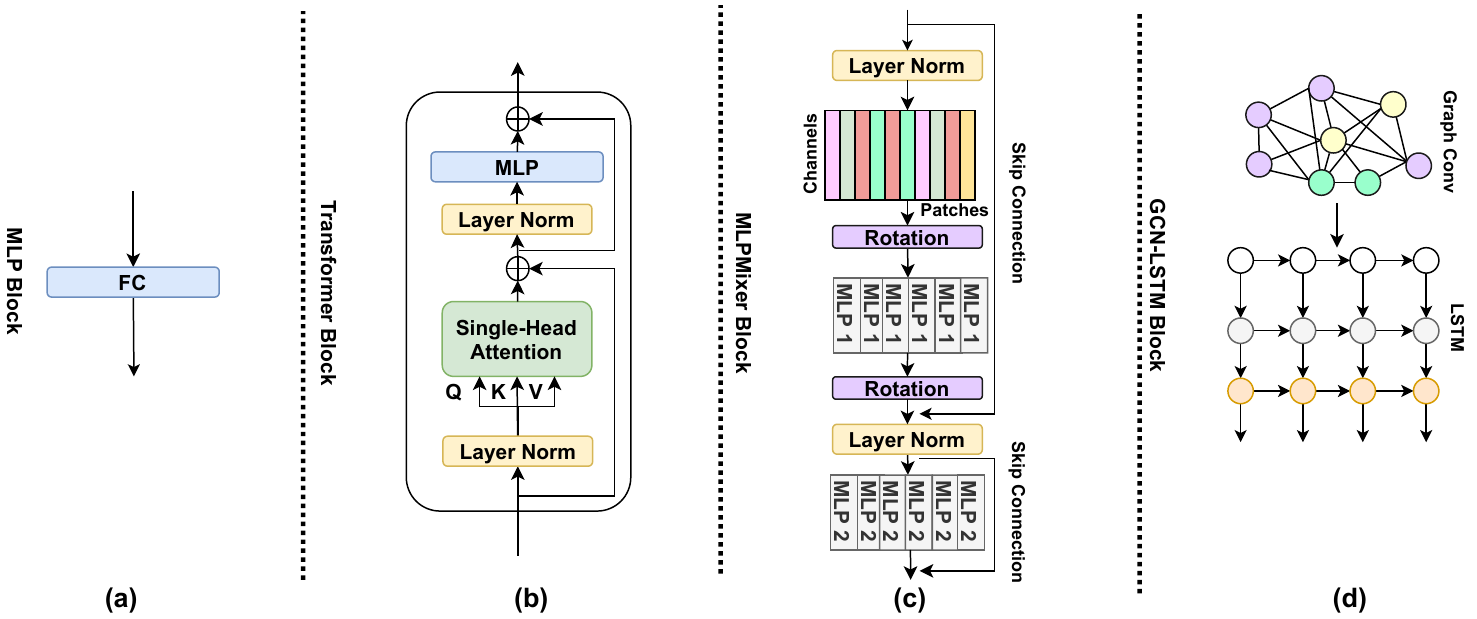}}
  \caption{Proposed simple neural-network baselines}
  \label{fig:NN_baselines}
\end{figure*}

\textbf{PCA reconstruction error:}
Our simplest reconstruction method can be seen as an outlier detection on a lower dimensional linear approximation of the train dataset single timestamp features.

After centering the training set $\mathbf{X}$ on its mean, using PCA, we compute the principal components of its features. 
This defines an affine approximation of $\mathbf{X}$ centered on the origin which can be expressed by the eigenvector matrix $\mathbf{U}\in\mathbb{R}^{F\cdot F'}$, where $F' < F$ is a fixed number of the first principle components. 
Then the test set $\hat{\mathbf{X}}$ is transformed to $\tilde{\mathbf{X}} = \hat{\mathbf{X}}\mathbf{U}^T\mathbf{U}\in\mathbb{R}^{\hat{T}\cdot F}$ and we consider the reconstruction error vectors $\mathbf{E} = err_{\mathbf{U}}(\hat{\mathbf{X}}) = \hat{\mathbf{X}} - \tilde{\mathbf{X}}$. 

There are two ways to interpret this transform. 
The first one is as a linear reconstruction of the test data, which is equivalent to using a linear autoencoder trained with the mean squared error loss on the training set, see \cite{pca-ae01} and \cite{pca-ae02}. 
The second way is to interpret it as the projection of each vector of $\hat{\mathbf{X}}$ to the linear subspace $\mathcal{S} = \text{span}(\text{cols}(\mathbf{U}))\subset\mathbb{R}^F$ formed by the principal components in $\mathbf{U}$. 
This interpretation highlights the linearity and simplicity of the method as each error vector $\mathbf{e}_t$ connects $\mathbf{x}_t$ with $\mathcal{S}$ and is perpendicular to $\mathcal{S}$, thus expresses the distance between $\mathbf{x}_t$ and $\mathcal{S}$. 

\subsection{Proposed neural network blocks baselines}
Contemporary anomaly detection techniques based on deep learning utilize modern neural networks to create solutions with varying levels of sophistication. 
Among the commonly employed architectures are auto-encoders (AE), long short-term memory (LSTM) networks, multi-layer perceptrons (MLPs), graph convolution networks (GCN), and Transformers. 
These neural network structures serve as the foundational components for designing intricate models intended for anomaly detection. 
In order to provide context for the usefulness of the more elaborate solutions, we utilize these architectures in their most basic form as a set of baselines. 
It is reasonable to expect that any solution which employs these as foundational components should perform better, provided they are trained on rich enough datasets of normal examples.
Our experiments demonstrate that, in most cases, these basic baselines perform better than models that incorporate a combination of these structures for the purpose of anomaly detection. 
Therefore, establishing such baselines may help understand the rationale behind the development of more complex models.

\textbf{1-layer linear MLP as auto-encoder:} As the first simplest neural baseline we use a single hidden-layer MLP without any activation as an auto-encoder.

\textbf{Single block MLP-Mixer:} Among the more modern variants of MLPs, the MLPMixer~\cite{mlp_mixer} has been shown to perform quite well on many vision problems. 
The architecture includes several MLP layers, called MLP-Mixer blocks. 
Each MLP-Mixer block consists of two sub-layers: a token-mixing sub-layer and a channel-mixing sub-layer. 
These operate on the spatial dimension and the channel dimension of the input feature maps.
The entire architecture consists of stacking several MLP-Mixer blocks, allowing the network to capture increasingly complex spatial and cross-channel dependencies in the input. 
We include a single standard block of MLPMixer as our baseline.

\textbf{Single Transformer block:} Since transformers are increasingly used in several recent anomaly detection methods, we use a basic transformer block with one single-head attention and one fully connected layer as a feed-forward output. 
This serves as the simplest and basic single transformer block baseline.

\textbf{1-layer GCN-LSTM block:} Using a single GCN layer feeding into a LSTM layer is a simple yet effective baseline for learning graph structure on multivariate time series data. 
The GCN layer is used to model the relationships between different time series variables, while the LSTM layer is used to capture temporal dependencies within each time series variable. 
The output of the LSTM layer is then forwarded to the output regression layer directly. Overall, this baseline provides a basic framework for jointly modeling the graph structure and temporal dependencies in multivariate time series data. 
Many recently published methods extend and improve upon this by incorporating additional GCN or LSTM layers, using attention mechanisms, or incorporating other types of graph neural networks.

Figure ~\ref{fig:NN_baselines} illustrates the proposed baseline neural network blocks. These baseline models are trained and compared in both reconstruction and forecasting modes. 

\subsection{Univariate time series representation}
Univariate time series data consist of a single observation at each timestamp, and most deep-learning methods designed for multivariate data are not directly applicable. 
Consequently, the most effective approaches for analyzing univariate data are typically focused on identifying unusual subsequences, or discords, within the time series. 
State-of-the-art discord discovery methods, for instance~\cite{Nakamura2020MERLINPD}, focus on optimizing the complexity and parameters of such methods that typically involve comparing windowed distances between timestamps. 
In this work, we use a similar yet effective representation for univariate time series data that allows the discovery of anomalies. 
Specifically, we represent each timestamp as a vector in $\mathbb{R}^{w+1}$, where $w$ denotes the number of preceding time stamps.
This representation can be efficiently computed in a sliding window fashion and has linear time complexity, making it efficient for practical use.
In section~\ref{subsec:abl-window}, we demonstrate that the impact of the window size on performance is relatively low and a small fixed window of $w=4$ suffice for the considered univariate datasets.

\subsection{Evaluation metrics}
\label{subsec:metrics}
A lot of papers introduced and criticised different metrics. 
In our view, anomaly detection shares a lot with object detection and semantic segmentation in computer vision, therefore it would need two metrics to fully capture model performance. 
The point-wise which captures the quality of the detection of individual anomalies and range-wise which expresses the quality of the anomaly segmentation. 
For the point-wise anomaly detection, we use the standard \textbf{F1 score}, which actually equals to the 1-dimensional Dice coefficient.
For completeness, we also include the flawed and commonly used \textbf{F1 score with point adjustment} denoted as $\mathbf{F1_{PA}}$.
For the range-wise metrics, we followed the work in this direction starting with the \textbf{Time-series precision and recall metrics} defined in \cite{tatbul2018precision} and then corrected for bias in \cite{wagner2023timesead} and we use the latter to compute an F1 score denoted as $\mathbf{F1_T}$.

Below are the definitions of the three scores we use together with the corresponding testing protocols:

$\mathbf{F1}$: Let $[\hat{y}_1,\ldots, \hat{y}_{\hat{T}}]$ be the ground truth per time-stamp on the test set and $[\tilde{y}_1^{thr},\ldots, \tilde{y}_{\hat{T}}^{thr}]$ the corresponding predictions set to 1 when $\tilde{y}_i > thr$ else to 0.
The hits are defined as $TP^{thr} = |\{i\leq \hat{T}\ |\ \tilde{y}_i^{thr}=\hat{y}_i\}|$, $FP^{thr} = |\{i\leq \hat{T}\ |\ \tilde{y}_i^{thr}=1\ and\ \hat{y}_i=0\}|$ and $FN^{thr} = |\{i\leq \hat{T}\ |\ \tilde{y}_i^{thr}=0\ and\ \hat{y}_i=1\}|$. Then the precision $Prec^{thr}$, recall $Rec^{thr}$ and F1-score $F1^{thr}$ are defined as usual based on those values. The final score is then $F1 = \max\limits_{thr\in \mathbb{R}} F1^{thr}$. 

$\mathbf{F1_{PA}}$: The final F1 score is computed exactly as before. This metric is different in its evaluation protocol which adjusts the predictions using the ground truth. Namely, for every contiguous anomaly interval $A = [t_1,\ldots, t_2]$ in the ground truth, if there is at least one $i\in A$ such that $\tilde{y}_i = 1$, then for every $j\in A$, $\tilde{y}_j$ is set to 1. In other words, if an anomaly interval is hit once by the predictions, then all predictions in the interval are corrected to match the ground truth. 

$\mathbf{F1_T}$:
Let $\mathcal{A}, \mathcal{P}$ be respectively the set of all ground truth and prediction anomaly intervals. Also let $\mathcal{P}_A = \{P\in\mathcal{P}\ |\ |A\cap P| > 0\}$ be the prediction intervals intersected by $A$. Then precision and recall are defined as follows:

$$Prec_T(\mathcal{A}, \mathcal{P}) = \frac{1}{|\mathcal{P}|}\sum\limits_{P\in\mathcal{P}} \gamma(|\mathcal{A}_P|, P) \frac{|\bigcup\mathcal{A}\cap P|}{|P|}$$

$$Rec_T(\mathcal{A}, \mathcal{P}) = \frac{1}{|\mathcal{A}|}\sum\limits_{A\in\mathcal{A}} \gamma(|\mathcal{P}_A|, A) \frac{|\bigcup\mathcal{P}\cap A|}{|A|}$$

The above definition is consistent with both \cite{tatbul2018precision} and \cite{wagner2023timesead}. The full formula in the latter paper for recall is 
$$Rec_T(\mathcal{A}, \mathcal{P}) = \frac{1}{|\mathcal{A}|}\sum\limits_{A\in\mathcal{A}} [\alpha \mathbbm{1}(|\mathcal{P}_A| > 0)$$
$$+ (1 - \alpha)\gamma(|\mathcal{P}_A|, A) \sum\limits_{P\in\mathcal{P}}\frac{\sum\limits_{t\in P\cap A} \delta(t - \min A, |A|)}{\sum\limits_{t\in A}\delta(t - \min A, |A|)}],$$
where $0\leq\alpha\leq 1$, $\delta\geq 1$ and $Prec_T(\mathcal{A}, \mathcal{P}) = Rec_T(\mathcal{P}, \mathcal{A})$. \cite{wagner2023timesead} proposed to fix the parameters $\alpha, \delta$ to $0$ and a constant function, in order to derive their formula for $\gamma$.

Under this assumption, we simplified those formulas to make them more comprehensible. Here we use the corrected $\gamma(n, A) = (\frac{|A| - 1}{|A|})^{n - 1}$ which guarantees that recall is increasing relative to the threshold of the anomaly detector. 
To provide some intuition, e.g. the recall computes an average of the fraction of ground truth intervals overlapped by the prediction which expresses the amount of discovery success. Every term is weighted though by $\gamma$ which decreases in value as multiple predictions hit the same ground truth interval, thus penalizing duplicates. Note that $Prec_T(\mathcal{A}, \mathcal{P}) = Rec_T(\mathcal{P}, \mathcal{A})$, i.e. precision measures the recall of prediction intervals by the ground truth. Finally, the F1-score denoted by $F1_T$ is defined as usual using $Prec_T$ and $Rec_T$.

The F1 scores are calculated using the best threshold computed on the test dataset and this threshold is also used to compute the corresponding precision and recall. Though we are not content with the threshold tuning, we choose this in order to follow the same protocol used in the published methods we have included for comparison. Here, it is important to also include the Area Under the
Precision Recall Curve (AUPRC) metric instead of only the F1 score obtained
with an optimal threshold. AUPRC provides a more realistic estimation of how well a method would perform in practical settings, where an estimated threshold based on a hold-out set would be used. In our appendix, we include tables (~\ref{tab:sota-standard},~\ref{tab:sota-wagner},~\ref{tab:ucr-standard},~\ref{tab:ucr-wagner}) with the separate precision, recall, and AUPRC values.

\section{Analysis}
\label{sec:experiments}

\textbf{Time series datasets:}
Overall, we used six commonly used benchmark datasets in our study.  Here, we report the details (Table~\ref{tab:dataset_sts}) and results from three multivariate datasets (SWaT, WADI, and SMD) and four univariate datasets (UCR/Internal Bleeding). The other two commonly used multivariate datasets (SMAP and MSL) have been identified in ~\cite{Wu_Keogh_2022} as potentially flawed containing trivial and unrealistic density of anomalies. For completeness, the descriptions and results of these two datasets are included in the appendix section~\ref{sec:supp_data}.

\begin{table}[t]
\scalebox{0.78}{
\begin{tabular}{lccccc}
\hline
Dataset  & Sensors (traces) & Train  & Test   & \#Anomalies (\%)\\
\hline
UCR/IB-16 & 1          & 1200   & 6301   & 12 (0.19\%)  \\
UCR/IB-17 & 1          & 1600   & 5900   & 111 (1.88\%)       \\
UCR/IB-18 & 1          & 2300   & 5200   & 102 (1.96\%)       \\
UCR/IB-19 & 1          & 3000   & 4500   & 10 (0 .22\%)        \\
\hline
{SWaT}                      & 51         & 47520  & 44991  & 4589 (12.20\%) \\
{WADI-127}                  & 127        & 118750      & 17280      & 1633 (9.45\%)                  \\
{WADI-112}                  & 112        & 118750 & 17280 & 918 (5.31\%) \\ 
{SMD}  & 38 (28)   & 25300 & 25300 & 1050 (4.21\%)  \\
\hline
\end{tabular}
}
\vspace{-0.05cm}
\caption{\small The statistical profile of the datasets in the experiment. }
\label{tab:dataset_sts}
\end{table}

Univariate HexagonML (UCR) datasets - InternalBleeding (IB)~\cite{Bert_ucr_internal_bleeding}: contains four univariate traces as the vital signs (arterial blood pressure). The anomalies are synthetic by adding a series of sine waves to one cycle or by injecting random numbers to a certain segment (Figure~\ref{fig:pca_ucr}). The unique and well-controlled anomalies in each trace allow a clean and sound evaluation among different approaches \cite{Wu_Keogh_2022}.

Secure Water Treatment (SWaT)~\cite{swat_dataset} and
Water Distribution (WADI)~\cite{wadi_dataset} datasets: contain sensor measurements of a water treatment test-bed.
Although SWaT is commonly used as a benchmark in recent publications, it should be noted that its use as a benchmark should be discontinued as it is flawed and unreliable \citeauthor{keogh_personal_comm} (personal communication, 7 May, 2024), see also~\cite{wagner2023timesead}.
The WADI dataset demonstrates the inconsistency in reporting performance comparisons in the TAD literature. The complete set of WADI contains 127 sensors (denoted as WADI-127 in our study). However, some recent methods~\cite{tuli2022tranad, gdn, kim2022TADevaluation, zekaietal-gta-2021, nsbf_kdd} use a specific subset of sensors when making comparisons without specifying the exact used sensors nor the reasons for such selection. Furthermore, in many cases, the selected subsets are inconsistent among competing methods.
In order to provide a fair overview of this impact on performance, we conducted our experiments on all the 127 WADI sensors (denoted as WADI-127) and on the subset of 112 sensors used in some recent studies~\cite{gdn} (denoted as WADI-112), separately.

Server Machine Dataset (SMD)~\cite{smd_dataset}:
contains 38 sensors from 28 machines for 10 days. Table~\ref{tab:dataset_sts} reports the average length of each trace.  Following the protocol, all models are trained on each machine separately and the  results are averaged from 28 different models.

\textbf{Evaluation:} 
We evaluate several state of the art representative deep learning based methods on commonly used timeseries benchmarks. To clearly show their utility, we evaluate these 1). under point-adjust $\mathbf{F1_{PA}}$ which is the common metric increasingly used in recent proposals. 2.) standard point-wise $\mathbf{F1}$ and 3.) Time-series range-wise metric $\mathbf{F1_T}$. See section~\ref{subsec:metrics} for the definitions. To highlight the prevalent use of flawed point-adjust $\mathbf{F1_{PA}}$, similar to~\cite{kim2022TADevaluation}, we also evaluate a random prediction:\\
\textbf{\color{red}Random:} The $\mathbf{F1_{PA}}$ protocol considers the whole interval of an anomaly as correctly predicted, as soon as the prediction considers a single point of the interval as anomalous. 
The random prediction directly shows that, under the point-adjust evaluation, methods might achieve high scores just because they have very noisy outputs. In
the random baseline setting, each timestamp is predicted
anomalous with probability 0.5 and we report the score achieved over five independent runs.

\begin{table*}[t]
\centering
\scalebox{0.75}{
\begin{tabular}{llrrrrrrrrrrrr} 
\hline
& & \multicolumn{3}{c}{\textbf{SWaT}} & \multicolumn{3}{c}{\textbf{WADI\_127}} & \multicolumn{3}{c}{\textbf{WADI\_112}} & \multicolumn{3}{c}{\textbf{SMD}}  \\
&  & \color{red}$F1_{PA}$ & $F1$ & $F1_{T}$ & \color{red}$F1_{PA}$ & $F1$ & $F1_{T}$ & \color{red}$F1_{PA}$ & $F1$ & $F1_{T}$ & \color{red}$F1_{PA}$ & $F1$ & $F1_{T}$ \\
\hline
\multirow{7}{*}{\rotatebox{90}{\parbox[inner-pos=b]{1.5cm}{\textbf{SOTA methods}}}} 
&  MERLIN~\cite{Nakamura2020MERLINPD} & 0.934 & 0.217 & 0.286 & 0.560 & 0.335 & 0.354 & 0.699 & 0.473 & 0.503 & 0.886 & 0.384 & 0.473 \\
&  DAGMM~\cite{dagmm} & 0.830 & 0.770 &  0.402 & 0.363 & 0.279 & 0.406 & \underline{0.829} & 0.520 & 0.609 & 0.840 & 0.435 & 0.379 \\
&  OmniAnomaly~\cite{omnianom} & 0.831 & 0.773 &  0.367 & 0.387 & 0.281 & 0.410 & 0.742 & 0.441 & 0.496 & 0.804 & 0.415 & 0.353 \\
&  USAD~\cite{usad} & 0.827 & 0.772 &  0.413 & 0.375 & 0.279 & 0.406 & 0.778 & 0.535 & 0.573 & 0.841 & 0.426 & 0.364 \\
&  GDN~\cite{gdn} & 0.866 & 0.810 &  0.385 & \underline{0.767} & 0.347 & 0.434 & \underline{0.833} & 0.571 & 0.588 & \textbf{0.929} & 0.526 & \underline{0.570} \\
&  TranAD~\cite{tuli2022tranad} & 0.865 & 0.799 & 0.425 & 0.671 & 0.340 & 0.353 & 0.680 & 0.511 & 0.589 & 0.827 & 0.457 & 0.390 \\
&  AnomalyTransformer~\cite{xu2022anomaly} &\underline{0.941} & 0.765 &  0.331 & 0.560 & 0.209 & 0.219 & 0.817 & 0.503 & 0.555 & 0.923 & 0.426 & 0.351 \\
\hline
\multirow{5}{*}{\rotatebox{90}{\parbox[inner-pos=b]{1.5cm}{\textbf{Simple baselines}}}} 
& \color{red}{Random} & \color{red}\textbf{0.963} & 0.218 & 0.217 & \color{red}\textbf{0.783} & 0.101 & 0.106 & \color{red}\textbf{0.907} & 0.101 & 0.106 & \color{red}0.894 & 0.080 & 0.080 \\
&  Sensor range deviation & 0.234 & 0.231 & 0.230 & 0.129 & 0.101 & 0.098 & 0.632 & 0.465 & 0.526 & 0.297 & 0.132 & 0.116 \\
&  L2-norm & 0.847 & 0.782 & 0.366 & 0.353 & 0.281 & 0.410 & 0.749 & 0.513 & 0.607 & 0.799 & 0.404 & 0.338 \\
&  1-NN distance & 0.847 & 0.782 & 0.372 & 0.372 & 0.281 & 0.410 & 0.751 & 0.568 & 0.618 & 0.833 & 0.463 & 0.384 \\
&  PCA Error & 0.895 & \textbf{0.833} & \textbf{0.574} & 0.621 & \textbf{0.501} & \textbf{0.557} & 0.783 & \textbf{0.655} & \textbf{0.699} & \underline{0.921} & \textbf{0.572} & \textbf{0.580} \\
\hline
\multirow{4}{*}{\rotatebox{90}{\parbox[inner-pos=c]{1cm}{\textbf{NN baselines}}}} 
& 1-Layer MLP & 0.856 & 0.771 & 0.519 & 0.295 & 0.267 & 0.384 & 0.601 & 0.502 & 0.558 & 0.829 & 0.514 & 0.487 \\
&  Single block MLPMixer & 0.865 & 0.780 & \underline{0.549} & 0.335 & 0.275 & 0.396 & 0.597 & 0.497 & 0.552 & 0.819 & 0.512 & 0.472 \\
&  Single Transformer block & 0.854 & 0.787 & 0.526 & 0.471 & 0.289 & 0.416 & 0.646 & 0.534 & 0.575 & 0.781 & 0.489 & 0.420 \\
&  1-Layer GCN-LSTM & 0.905 & \underline{0.829} & 0.532 & 0.593 & \underline{0.439} & \underline{0.540} & 0.748 & \underline{0.596} & \underline{0.645} & 0.847 & \underline{0.550} & 0.535 \\
\hline
\end{tabular}
}
\caption{\small Experimental results for SWaT, WADI, and SMD datasets. The bold and underline marks the best and second-best value. $F1_{PA}$: F1 score with point-adjust; $F1$: the standard point-wise F1 score; $F1_{T}$: time-series range-wise F1 score}
\label{tab:results}
\end{table*}

\subsection{Model setup}
In this section we summarize our data preprocessing steps and the hyperparameters used to train the models. 
The features were scaled to the interval $[0, 1]$ in the training dataset and the learned scaling parameters were used to scale the testing dataset.
For all of our NN baselines, when trained in forecasting mode, we used a time window of size 5.  
We used a 90/10 split to make the train and the validation set. The validation set is only used for early stopping to avoid over-fitting and the Adam optimizer with learning rate 0.001 and a batch size of 512 were used.

\textbf{PCA reconstruction error}: For multivariate data, this method uses the first 30 principal components when data has more than 50 sensors and 10 otherwise. On univariate datasets, the first 2 principal components with a window size of 5 are used.

\textbf{1-layer Linear MLP}: A hidden layer of size 32 is used.

\textbf{Single block MLP-Mixer} and \textbf{Single Transformer block} both use an embedding of 128 for the hidden layer.

\textbf{1-layer GCN-LSTM block}: The dimension for the GCN output nodes is set to 10 and for LSTM layer to 64 units.

Our neural network baselines are trained in the forecasting mode, similar to most other methods we are comparing with. We also provide their performance for the reconstruction mode in the appendix section~\ref{subsec:abl-reco}.

\textbf{Hyperparameter sensitivity:} Most of the simple baselines don't have tunable hyperparameters. The only exceptions are the projection dimension of the PCA method and the sliding window for univariate series. We have included their ablations in sections~\ref{subsec:abl-pca} and ~\ref{subsec:abl-window}. We trained our neural network baseline models using the same hyperparameters as stated above on all multivariate datasets. The purpose of this analysis was to demonstrate that even with basic hyperparameters, these simple neural networks can achieve comparable performance to SOTA deep learning models. The fact that the hyperparameters of the SOTA models were optimized for each respective datasets, while the simple NN baseline models used the same set of hyperparameters, highlights less reliance on dataset-specific tuning.

\textbf{Published SOTA methods}: All methods were trained with the hyper-parameters recommended in their respective papers, where possible, with their official implementations or the implementations provided in~\cite{tuli2022tranad}. GDN~\cite{gdn} on WADI-112 is not re-trained since the authors provided the trained checkpoint of their official model.

\subsection{Model performance overview}
Table~\ref{tab:results} outlines the model performance on the three multivariate benchmark datasets, SWaT, WADI, and SMD.

First, it is evident that all methods have higher scores on the predominantly used point-adjusted $F1_{PA}$ metric including the \textit{random prediction} which performs better in almost all comparisons. 
This artificial advantage created by point-adjust is not present on the pure F1 score protocols which do not favour noisy random predictions.
On both standard point-wise $F1$ and range-wise $F1_{T}$ metrics, the simple baselines such as PCA reconstruction error performs better on all datasets while other baselines such as 1-NN distance and L2-norm are often very close to the best performing methods.
Furthermore, the NN-baselines in most cases outperform the more complex SOTA deep models which are build using these as basic building blocks.
This is a strong evidence that the complicated solutions introduced to solve the TAD task do not provide a benefit compared to such simple baselines.
Finally, one can notice the interplay between the point-wise and range-wise metrics.
In datasets like SWAT, where there is a small number of long anomaly intervals, the $F1$ score is much higher than the $F1_T$ score, on noisy datasets with more consistent anomaly lengths, like WADI\_127, $F1_T$ is tendentially higher, while on cleaner datasets with frequent short anomalies, like univariate UCR datasets, the two scores are comparable. 

\begin{table*}[t]
\centering
\scalebox{0.76}{
\begin{tabular}{lrrrrrrrrrrrrr}
\hline
& & \multicolumn{3}{c}{\textbf{UCR/IB-16}} & \multicolumn{3}{c}{\textbf{UCR/IB-17}} & \multicolumn{3}{c}{\textbf{UCR/IB-18}} & \multicolumn{3}{c}{\textbf{UCR/IB-19}} \\
& & $F1_{PA}$ & $F1$ & $F1_T$ & $F1_{PA}$ & $F1$ & $F1_T$ & $F1_{PA}$ & $F1$ & $F1_T$ & $F1_{PA}$ & $F1$ & $F1_T$ \\
\hline
& LOF~\cite{breunig2000lof}	& 0.878	& 0.476	& 0.476	& 1.000	& 0.959	& 0.955	& 1.000	& 0.915& 0.911	& 1.000	& 0.857	& 0.857 \\

& MERLIN~\cite{Nakamura2020MERLINPD}	& 1.000	& \textbf{0.846}	& \textbf{0.846}	& 1.000	& \textbf{0.987}	& \textbf{0.987}	& 1.000	& 0.795	& 0.795	& 1.000	& \underline{0.870}	& \underline{0.870} \\
\hline
\multirow{5}{*}{\rotatebox{90}{\parbox[inner-pos=b]{1.5cm}{\textbf{Simple baselines}}}}
& Random	& 0.151	& 0.005	& 0.030	& 0.941	& 0.041	& 0.116	& 0.887	& 0.039	& 0.039	& 0.488	& 0.030 & 0.030 \\
& Sensor range deviation	& 0.000	& 0.003	& 0.000	& 0.902	& 0.085	& 0.094	& 0.000	& 0.038 & 0.038	& 0.000	& 0.004	& 0.004 \\
& L2-norm	& 0.014	& 0.011 & 0.021	& 0.276	& 0.058	& 0.164	& 0.241	& 0.061	& 0.061	& 0.028 & 0.017	& 0.017 \\
& 1-NN distance	& 0.828	& \underline{0.786}	& \underline{0.786}	& 1.000	& 0.973	& 0.969	& 1.000	& \underline{0.889}	& \underline{0.889}	& 1.000	& \underline{0.870}	& \underline{0.870} \\
& PCA Error	& 0.889	& 0.750	& 0.750	& 1.000	& \underline{0.974}	& \underline{0.974}	& 1.000	& \textbf{0.990}	& \textbf{0.990} &  1.000	& \textbf{1.000}	& \textbf{1.000} \\
\hline
\end{tabular}
}
\caption{\small Comparison of simple baselines on four univariate UCR/InternalBleeding datasets.
}
\label{tab:ucr-result-rebuttal}
\end{table*}

\begin{figure}[t]
\includegraphics[width=0.474\textwidth]{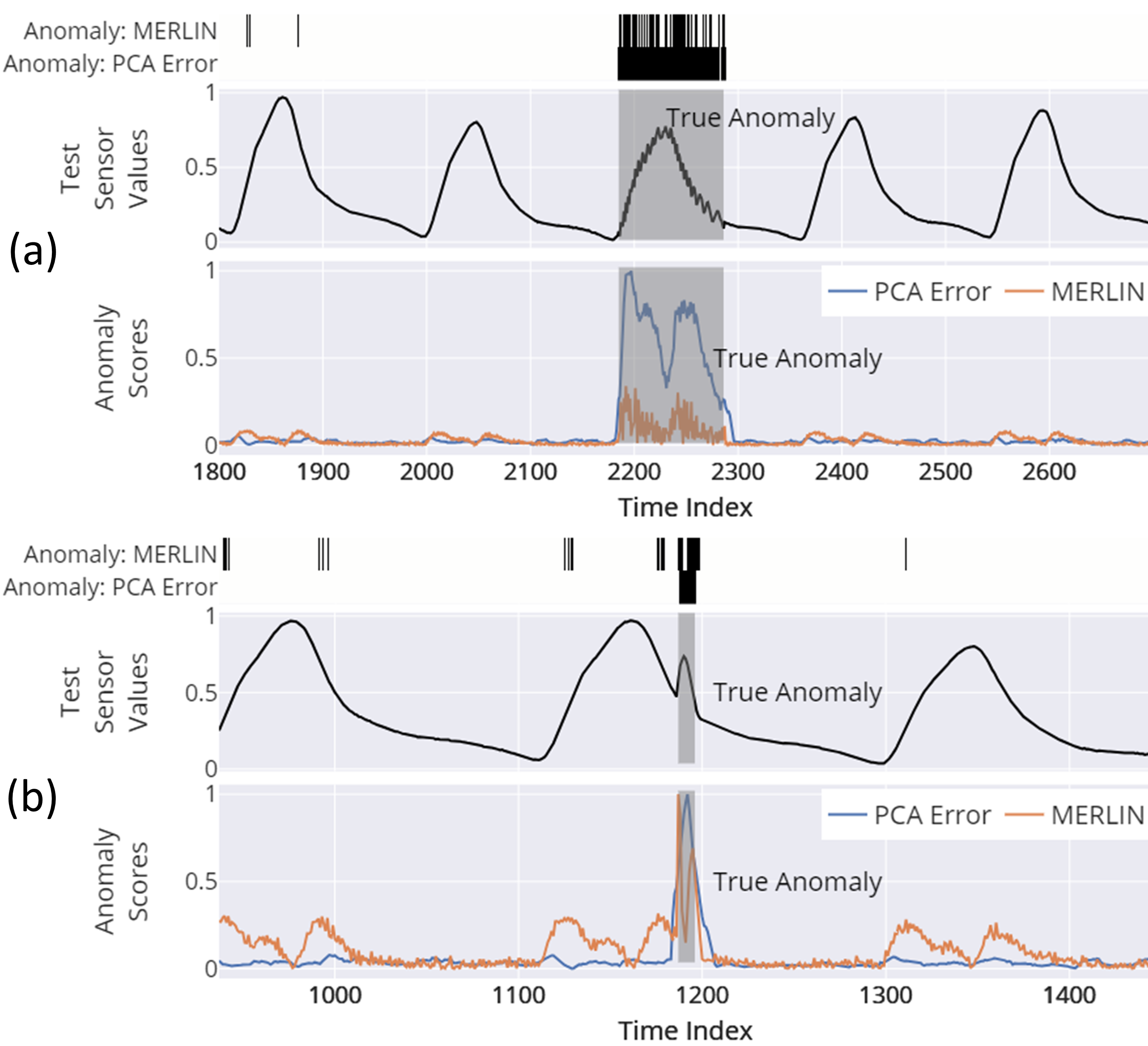}
\caption{\small 
Visual comparison: 
The gray shaded areas denote the ground truth anomalies. 
(a) UCR/IB-18 dataset with a series of sine waves added as anomaly. 
(b) UCR/IB-19 dataset with random numbers added as anomaly. 
}
\label{fig:pca_ucr}
\end{figure}

Table~\ref{tab:ucr-result-rebuttal} provides a comparison on univariate UCR datasets with our simple baselines. Here we include two representative univariate TAD methods, a highly effective classic method Local Outlier Factor (LOF)~\cite{breunig2000lof} and a more recent SOTA method Merlin~\cite{Nakamura2020MERLINPD}.
As shown in Figure~\ref{fig:pca_ucr} the normal periodical phase-shift and magnitude changes, which are considered normal in the light of physiology, are misclassified as anomalies by such methods in contrast to the simple PCA-Error baseline.

\subsection{Analysis of the deep models learned function}
The consistently better results of the simple methods raises the question of what type of functions are learned by the more complicated deep learning models. To investigate this, we try to approximate the behavior of the most prominent of the deep learning models by linear functions. We achieve this by performing a simple form of distillation. Given a deep learning model $M_{\theta}$ trained on the training data $\mathbf{X}$, we compute its predictions $M(\mathbf{X})\subset\mathbb{R}^F$ and then train a linear model $L$ on the data/target tuple $(\mathbf{X}, M(\mathbf{X}))$ using a mean squared error (MSE) loss. The linear model in this case is simply a 1-layer perceptron. Upon evaluating both $M$ and $L$ on the test set on the anomaly detection task, we observed that their scores are very close and they exhibited high agreement on their predictions. Table~\ref{tab:linear_fits} depicts this with the linear model $L$ marked as `Line' and the corresponding deep learning model $M$ marked as `Orig'. The performance of distilled linear version of the complex models suggests that even though the learned functions may be complex and may improve forecasting, their ability to distinguish anomalies can still be effectively captured by linearizing them.

\begin{table}[t]
\scalebox{0.85}{
\begin{tabular}{lllllllll}
\hline
\multicolumn{1}{c}{Methods} & \multicolumn{2}{c}{\textbf{SWaT}} & \multicolumn{2}{l}{\textbf{WADI\_112}} \\
\multicolumn{1}{c}{}  & Orig & Line & Orig & Line \\
\hline
Single block MLPMixer& 0.780 & 0.770 & 0.497 & 0.500 \\ 
Single Transformer block & 0.787 & 0.772 & 0.534 & 0.521 \\ 
1-Layer GCN-LSTM  &  0.829 & 0.794 & 0.596 & 0.587 \\ 
\hline
TranAD~\cite{tuli2022tranad} &   0.799    &  0.800     &    0.511   &   0.572    \\
GDN~\cite{gdn} &   0.810    &     0.808  &    0.571   &  0.543     \\ 
\hline
\end{tabular}
}
\caption{\small Linear approximation of complex models on two datasets. \textbf{Orig}: original model \textbf{Line}: linear approximated mode. Performance is reported on the standard point-wise $F1$ score.}
\label{tab:linear_fits}
\end{table}


\begin{table*}[t]
\centering
\scalebox{0.75}{
\begin{tabular}{crrrrrrr|rrrrrrr} 
\hline
{} & & \multicolumn{6}{c}{\textbf{SWAT}} &  \multicolumn{6}{c}{\textbf{WADI-112}}  \\
\cline{4-7}  \cline{10-13} \\ 
{} & & \multicolumn{2}{c}{\textbf{\underline{None}}} &
\multicolumn{2}{c}{\textbf{\underline{Mean-STD}}} &
\multicolumn{2}{c}{\textbf{\underline{Median-IQR}}} &
\multicolumn{2}{c}{\textbf{\underline{None}}} &
\multicolumn{2}{c}{\textbf{\underline{Mean-STD}}}&
\multicolumn{2}{c}{\textbf{\underline{Median-IQR}}}
\\ 
\hline
{} &  &  F1 & AUPRC & F1 & AUPRC & F1 & AUPRC & F1 & AUPRC & F1 & AUPRC & F1 & AUPRC \\
\hline
& GDN~\cite{gdn} & 0.767& 0.724 & 0.685 & 0.473 & 0.810 & 0.762 & 0.549& 0.492& 0.456 & 0.353 & 0.571 & 0.519 \\

& TranAD~\cite{tuli2022tranad} & 0.769& 0.708 & 0.742 & 0.638 & 0.799 & 0.764 & 0.511& 0.529& 0.448 & 0.312 & 0.509 & 0.554 \\

& PCA Error & 0.802 & 0.725 & 0.833 & 0.744 & 0.756 & 0.721 & 0.600 & 0.591 & 0.513 & 0.351 & 0.654 & 0.570   \\

& 1-Layer GCN-LSTM & 0.770 & 0.715 & 0.775 & 0.660 & 0.829 & 0.792 & 0.592 & 0.520 & 0.520 & 0.411 & 0.596 & 0.535 \\
\hline
\end{tabular}
}
\caption{\small Impact of normalization on scores. Normalisation of prediction scores before thresholding impacts performance. Performance is reported on the point-wise $F1$ and $AUPRC$ score.
}
\label{tab:norm_results}
\end{table*}

\subsection{Ablation: Impact of normalization} 
\label{subsec:abl-norm}

Anomaly detection methods for multivariate datasets often employ normalization and smoothing techniques to address abrupt changes in prediction scores that are not accurately predicted. However, the choice of normalization method before thresholding can impact performance on different datasets. In Table~\ref{tab:norm_results}, we compared the performance with and without normalization. We consider two normalization methods, mean-standard deviation and median-IQR, on two datasets. Our analysis shows that median-IQR normalization, which is also utilized in the GDN~\cite{gdn} method, improves performance on noisier datasets such as WADI. In Table \ref{tab:results}, we have presented the best performance achieved by each method, including our baselines and considered state-of-the-art models, using either none or one of these normalisation, whichever is applicable.

\subsection{Ablation: PCA Error projection dimension}\label{subsec:abl-pca}
\begin{figure}[t]
  \centerline{\includegraphics[width=0.45\textwidth]{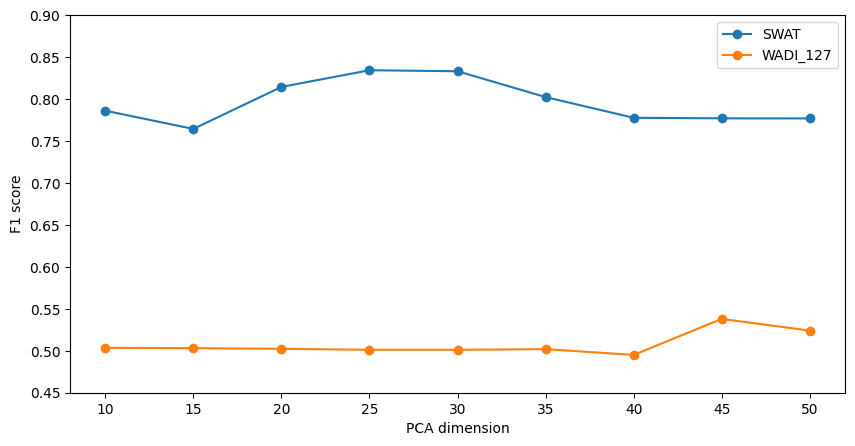}}  
  \caption{\small Point-wise F1 score as a function of the PCA dimension for the PCA Error method, evaluated on the SWAT and WADI\_127 datasets.}
  \label{fig:pca_dim}
\end{figure}

On all the multivariate datasets with more than 50 sensors (i.e., SWaT and WADI) our PCA Error baseline approach utilized the first 30 eigenvectors for the PCA projection. In Figure~\ref{fig:pca_dim}, we present the performance as a function of varying PCA projection dimensions. It is observed that higher projection dimensions may be more beneficial for WADI (127-dimensional) compared to SWaT (51-dimensional). However, the optimal projection dimension should be determined using a validation set as it may impact performance.  Unlike more sophisticated techniques with several hyperparameters specifically configured for each dataset, the baseline approach of using PCA with a fixed number of eigenvectors is relatively simple and easily tunable. 

\section{Quo vadis}

As we have demonstrated, a plethora of deep learning approaches introduced to solve the task of TAD were outperformed by simple neural networks and linear baselines. 
Furthermore, when distilling some of those methods to linear models, their performance remained almost unchanged. 
There could be several causes for this issue for example the over-fitting on the normal data or the existence of too high aleatoric uncertainty which makes it hard to separate the difficult anomalies from normal sections. 
In any case, the main takeaway is that those methods, though potentially useful for other time-series tasks such as forecasting, do not bring much additional value for the task of TAD and their complexity is definitely not justified. 
What is even more worrisome, is that they managed to create up to now an illusion of progress due to the use of a flawed evaluation protocol, inadequate metrics and the lack or low quality of benchmarking with simpler methods. 

We cannot stress enough the fact that almost all the recent deep-learning based methods use the point-adjust post-processing step often \textit{without clearly stating} this. 
Under this evaluation these models implicitly optimize for near random predictions where their high performance is used as evidence of their proposed model's utility. 
An example of this trend presented at recent leading Machine Learning venues is \cite{xu2022anomaly, puad_icml, zhou2023one}.
Another common malpractice, is the use of mismatched evaluation metrics in tables i.e., applying point-adjust and directly comparing their results to other methods which were scored without it. 
Similar issues are observed in dataset discrepancies like the introduction of new versions of a dataset which use a subset of the sensors and result in higher scores.

Aside from exposing the limitations of these methods, we provide a comprehensive set of simple benchmarks which can help re-start investigations in TAD starting on a solid baseline.
We think that those methods will pinpoint which anomalies are easy to detect and which ones are the challenging ones that should be detected if any progress is to be made. 
This is further reinforced by the fact that there seem to be a high agreement between detected and undetected anomalies between all methods investigated. 
We provide an analysis of this agreement in the appendix section~\ref{sec:model_analysis}. 
This agreement leads us to believe that the current datasets used in TAD are, in some sense, simultaneously too hard and too easy. 
The fact that so many complex deep learning architectures have been developed to tackle the hard anomalies in those datasets, but failed, is unsatisfactory, but maybe not unexpected.
More comprehensive datasets with a spread spectrum of difficulty in anomalies could provide an incremental improvement path and means of properly comparing methods. 

Furthermore, we believe that evaluation using both point-wise and range-wise methods will help better compare methods and identify their strengths and weaknesses.

We hope our work will help improve the research efforts on TAD by triggering focus on the introduction of new and richer datasets, increasing awareness of limitations of current evaluation protocols, and encouraging caution in the premature adoption of complex tools for the task.

\section*{Impact Statement}
"This paper presents work whose goal is to advance the field of Machine Learning. There are many potential societal consequences of our work, none which we feel must be specifically highlighted here.”
\bibliography{icml_manuscript}
\bibliographystyle{icml2024}

\newpage
\appendix
\onecolumn

\section{Appendix}
\label{sec:appendix}

In the following appendix, we present several analyses and ablation studies related to the results discussed in the main paper. It is structured as follows:
\begin{enumerate}
    \item \textit{Analysis}: We analyze the agreement on the detected anomalies between the different models (Figures~\ref{fig:swat_agreements} and~\ref{fig:wadi_agreements}).
    \item \textit{Additional evaluations/ablations}: Several studies are presented related to the evaluation of the model performances:
    \begin{itemize}
        \item \textit{Ablation window size for Univariate data}: We show the impact of the sliding window size on the performance of our simple baselines on univariate data (Figure~\ref{fig:window_dim}).
        \item \textit{NN-baselines: reconstruction vs forecasting mode}: We show the performance of our neural network baselines when trained in reconstruction and forecasting mode (Table~\ref{tab:nn_recons}).
        \item \textit{Detailed performance comparison}: At the end, we include 
        detailed tables (Table~\ref{tab:sota-standard}, Table~\ref{tab:sota-wagner}, Table~\ref{tab:ucr-standard}, Table~\ref{tab:ucr-wagner}) with performance comparison of all methods reporting their F1, precision, recall and AUPRC under both standard point-wise and time-series range-wise metrics.
    \end{itemize}
    \item \textit{Performance of our simple baselines on SMAP and MSL datsets}: We include a comparison of our simple baseline methods and various SOTA methods on the additional multivariate SMAP and MSL datasets (Table~\ref{tab:simple_baseline_smap_msl_smd}).
\end{enumerate}

\subsection{Analysis of model agreement on the detected anomalies} \label{sec:model_analysis}

We have noticed a very high agreement on the anomalies detected by the different methods. Those agreements are especially pronounced between the SOTA deep learning methods. In order to quantify them, we compute a score similar to mAP in object detection which measures the agreement between two different predictions restricted to the ground truth anomaly intervals. The score is defined as follows:

\begin{figure*}[!tbh]
\begin{subfigure}[c]{0.45\linewidth}
\centering
\includegraphics[width=1\linewidth]{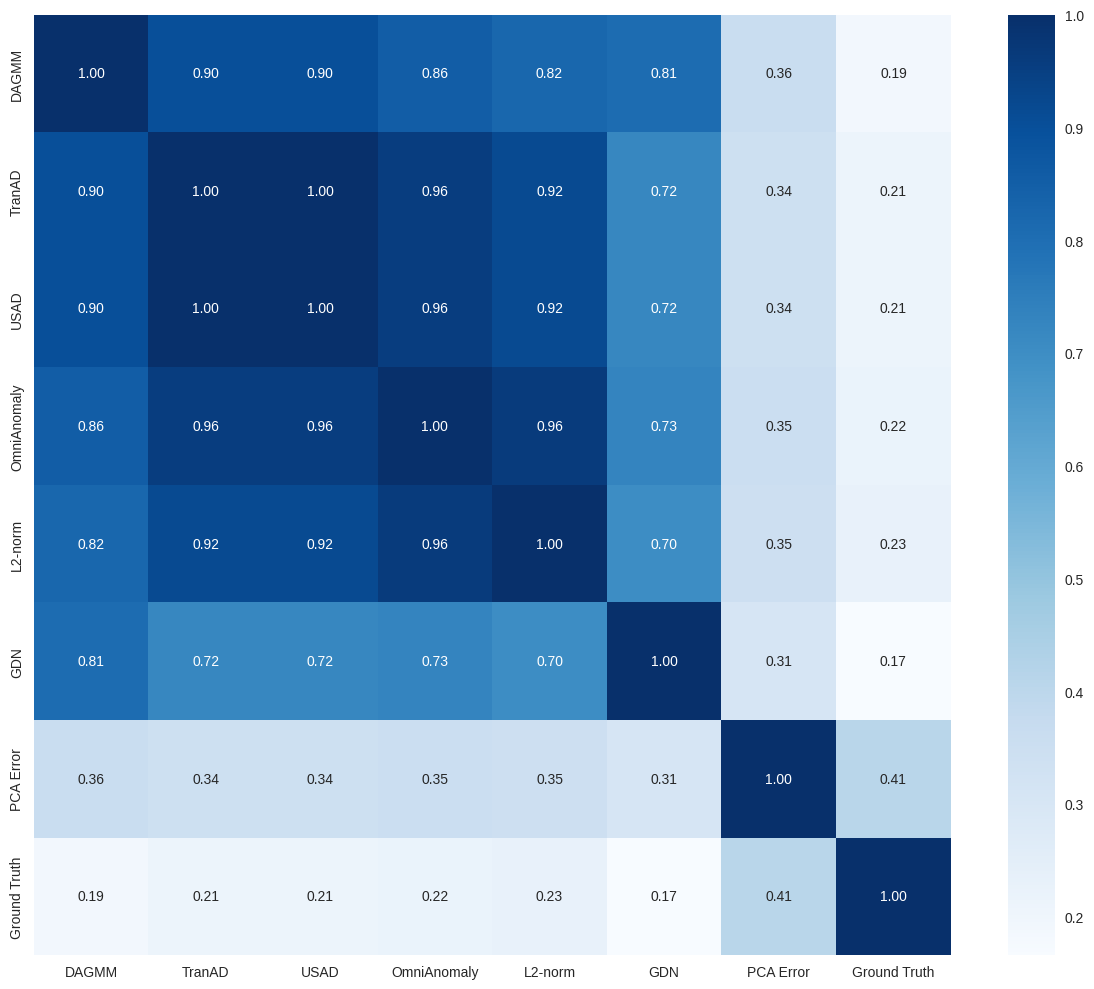}
\caption{SWAT agreement matrix between methods expressed as the IOU of the sets of interval indices averaged over the hit ratio thresholds in $[0.2:0.95:0.05]$.}
\label{fig:swat_agreements}
\end{subfigure}
\hfill
\begin{subfigure}[c]{0.45\linewidth}
\centering
\includegraphics[width=1\linewidth]{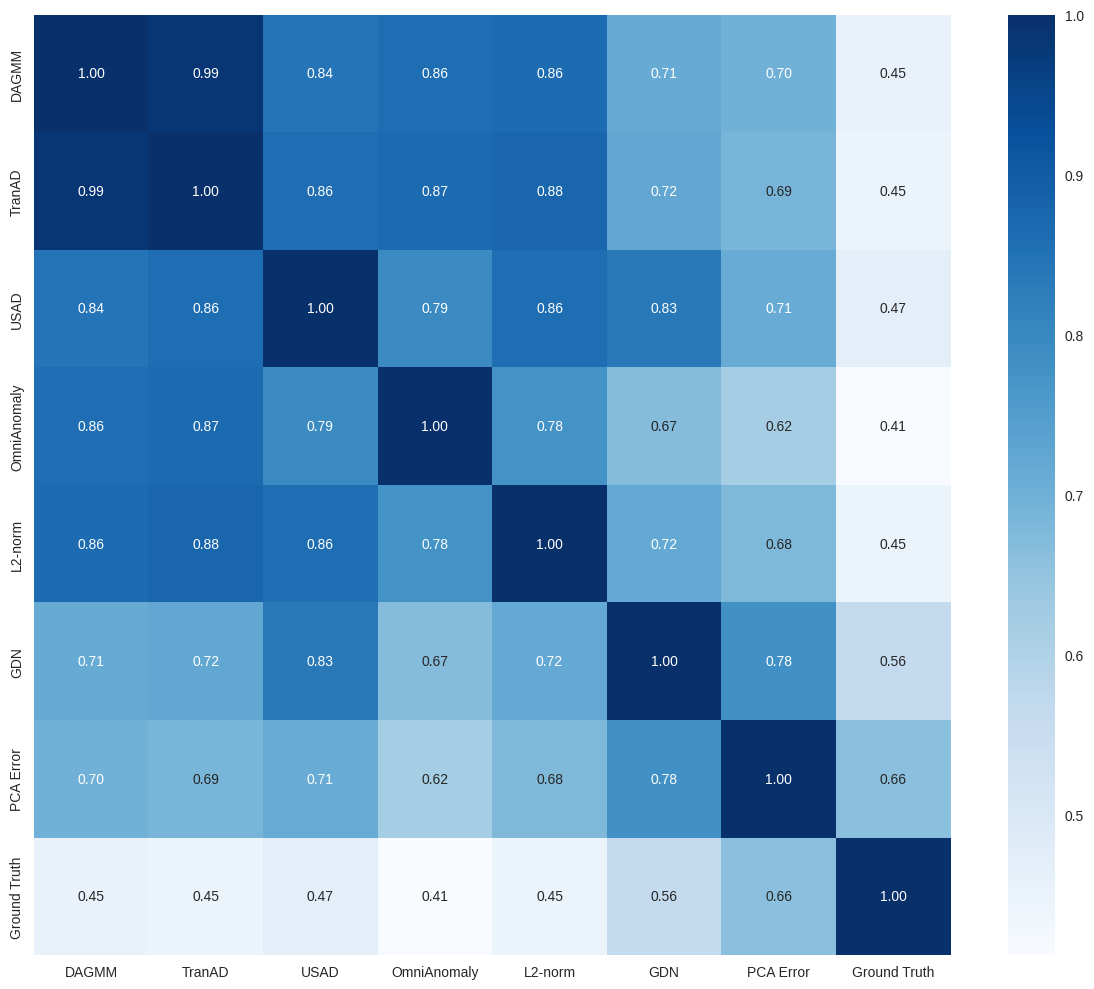}
\caption{WADI\_112 agreement matrix between methods expressed as the IOU of the sets of interval indices averaged over the hit ratio thresholds in $[0.2:0.95:0.05]$.}
\label{fig:wadi_agreements}
\end{subfigure}
\caption{Analysis of model agreement on the detected
anomalies}
\end{figure*}

Assume $A = \{[a_1, b_1],\ldots [a_K, b_K]\}$ are the $K$ ground truth anomaly intervals, defined by their start and end timestamp indices as integer intervals. Thus for the interval on index $s$, $\hat{y}_i=1$ for all $t\in[a_s, b_s]$. For an interval $[a_s, b_s]$ and a prediction $\tilde{y}$, the hit ratio is the ratio $\frac{|\{t\in[a_s, b_s]: \tilde{y}_t = 1\}|}{|[a_s, b_s]|}$ of the timestamps with a positive prediction in $[a_s, b_s]$ to the total number of timestamps in $[a_s, b_s]$. For a given prediction $\tilde{y}$ and a hit ratio threshold $r$, the detected anomaly intervals is the index list $H_{\tilde{y}} = \{i_1,\ldots i_L\}\subseteq [1, L]$ of intervals for which the prediction has hit ratio above $r$. For two different predictions $\tilde{y}^1$ and $\tilde{y}^2$, the agreement between them on a given hit ratio threshold $r$ is defined as the intersection over union (IOU) of the index sets $H_{\tilde{y}^1}, H_{\tilde{y}^2}$ for $r$. Finally, the average agreement between two predictions $\tilde{y}^1$ and $\tilde{y}^2$ is the mean of their agreements over all the thresholds from 0.2 to 0.95 with step 0.05.

In Figures~\ref{fig:swat_agreements} and~\ref{fig:wadi_agreements} the matrices of the agreements between all models and the ground truth are displayed for the SWAT and WADI-112 datasets. In both cases, the agreement between different models is much higher compared to the agreement to the ground truth, indicating that the models learn to recognize similar anomalies. Only the GDN model and even more the PCA Error baseline seem to have a comparably higher agreement with the ground truth.


\subsection{Additional evaluations/ablations}\label{sec:abl-eval}

\subsubsection{Ablation window size for Univariate data}\label{subsec:abl-window}
\begin{figure}[t]
  \centerline{\includegraphics[width=0.45\textwidth]{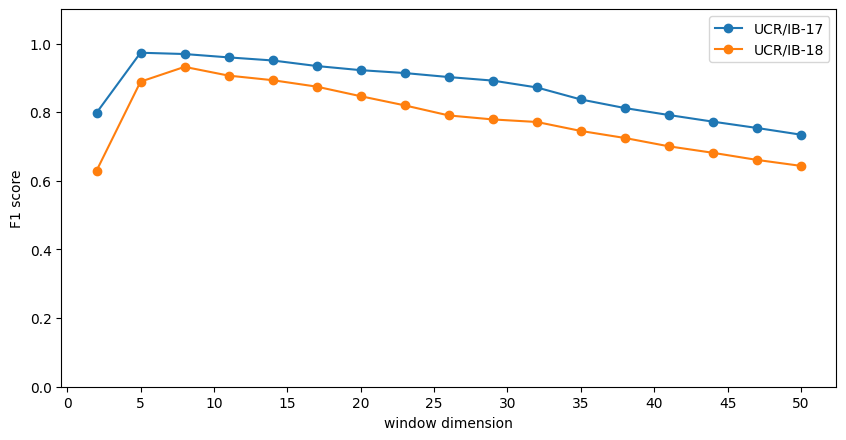}}  
  \caption{Impact of sliding window size to generate univariate data representation on the two UCR dataset traces UCR/IB-17 and UCR/IB-18.}
  \label{fig:window_dim}
\end{figure}

As outlined in section 3.2 of the main paper, we created an effective univariate data representation by concatenating past observations with the current timestamp using a sliding window approach. We discovered that this basic representation yielded effective results with a window size of $w=4$ leading to a 5-dimensional representation space. Figure~\ref{fig:window_dim} displays the performance impact based on the window size. This plot illustrates that a smaller window over 4-5 past observations is a reasonable choice for the UCR datasets, while larger window dimensions do not add any further advantage. We opted to use our simple 1-NN distance approach and varied the window sizes to avoid manipulating any other parameters.

\subsubsection{NN-baselines: reconstruction vs forecasting mode}\label{subsec:abl-reco}
In our main paper, we demonstrated the effectiveness of our simple neural network baselines when trained in forecasting mode, which is in line with most state-of-the-art deep learning models we compared with. During training, the output before the final target dense regression layer has a shape of (batch-size, sequence, embedding-dim). In forecasting mode, we use a 1-D global average pooling to project it to (batch-size, 1, embedding-dim). However, we can skip the average-pooling operation and train these models in a reconstruction (auto-encoding) fashion. For completeness, we present their performance in reconstruction mode in Table~\ref{tab:nn_recons}. Our results show that the performance of these models in reconstruction mode is comparable to that in forecasting mode, particularly considering the impact of random seed between training runs. Therefore, there does not appear to be any significant advantage in training these models in forecasting mode, at least for the datasets we considered.

\begin{table*}[t]
\centering
\scalebox{0.8}{
\begin{tabular}{lllllllllllll} 
\hline
\multicolumn{1}{c}{\multirow{3}{*}{\textbf{Method }}} & \multicolumn{6}{c}{\textbf{Reconstruction}} & \multicolumn{6}{c}{\textbf{Forecasting}} \\
\multicolumn{1}{c}{} & \multicolumn{2}{c}{\textbf{SWAT}} & \multicolumn{2}{l}{\textbf{WADI 127}} & \multicolumn{2}{c}{\textbf{WADI 112 }} & \multicolumn{2}{c}{\textbf{SWAT}} & \multicolumn{2}{c}{\textbf{WADI 127}} & \multicolumn{2}{c}{\textbf{WADI 112}} \\
\multicolumn{1}{c}{} & \multicolumn{1}{c}{\textbf{F1}} & \multicolumn{1}{c}{\textbf{AUPRC}} & \multicolumn{1}{c}{\textbf{F1}} & \multicolumn{1}{c}{\textbf{AUPRC}} & \multicolumn{1}{c}{\textbf{F1}} & \multicolumn{1}{c}{\textbf{AUPRC}} & \multicolumn{1}{c}{\textbf{F1}} & \multicolumn{1}{c}{\textbf{AUPRC}} & \multicolumn{1}{c}{\textbf{F1}} & \multicolumn{1}{c}{\textbf{AUPRC}} & \multicolumn{1}{c}{\textbf{F1}} & \multicolumn{1}{c}{\textbf{AUPRC}} \\ 
\hline
1-Layer MLP & 0.771 & 0.707 & 0.161 & 0.139 & 0.580 & 0.517 & 0.771 & 0.797 & 0.267 & 0.193 & 0.501 & 0.437 \\
Single block MLPmixer & 0.770 & 0.706 & 0.289 & 0.210 & 0.530 & 0.475 & 0.779 & 0.791 & 0.274 & 0.217 & 0.496 & 0.426 \\
Single Transformer Block & 0.770 & 0.702 & 0.448 & 0.410 & 0.527 & 0.798 & 0.795 & 0.874 & 0.319 & 0.662 & 0.538 & 0.756 \\
1-Layer GCN-LSTM & 0.770 & 0.714 & 0.498 & 0.451 & 0.577 & 0.486 & 0.829 & 0.792 & 0.439 & 0.367 & 0.596 & 0.535 \\
\hline
\end{tabular} }
 \caption{NN-Baselines: Reconstruction vs Forecasting.}
\label{tab:nn_recons}
\end{table*}

\subsubsection{Detailed performance comparison}
Finally, we provide tables which contain the detailed scores of all models in terms of precision, recall, F1-score and area under the precision-recall curve (AUPRC). For the multivariate time series datasets, Table~\ref{tab:sota-standard} shows the evaluation under point-wise metrics; Table~\ref{tab:sota-wagner} shows the evaluation under time series range-wise metrics \cite{wagner2023timesead}. Similarly, for the univariate datasets, Table~\ref{tab:ucr-standard} evaluates under point-wise and Table~\ref{tab:ucr-wagner} provides the performance under range-wise metrics.

\subsection{Performance of our simple baselines on SMAP and MSL datasets} \label{sec:supp_data}

Soil Moisture Active Passive (SMAP) and Mars Science Laboratory (MSL) datasets, collected from a spacecraft of NASA~\cite{smap_msl_datasets}, are another two widely utilized benchmark datasets in the literature. The SMAP dataset contains information on soil samples and telemetry of the Mars rover; the MSL dataset comes from the actuator and sensor data for the Mars rover itself. Although these benchmark datasets are widely used in the literature, their quality and validity suffer from several pitfalls, such as triviality, mislabeling, and unrealistic density of anomaly (see \citeauthor{Wu_Keogh_2022} (\citeyear{Wu_Keogh_2022}) for details). The statistics profile of each dataset is listed in Table~\ref{tab:appendix_dataset_sts}. Since each dataset contains traces with various lengths in both the training and test sets, we report the average length of traces and the average number of anomalies among all traces per dataset. We also report the total number of data points and anomalies per dataset for the clarity of comparison in the literature. 

\begin{table*}[h]
\centering
\scalebox{0.95}{
\begin{tabular}{lcccccc}
\hline
Dataset  & No. Sensors (Traces) & Avg. Train (Total) & Avg. Test (Total)  & Avg. Anomalies (\%) & Total Anomalies (\%)\\
\hline
MSL  & 55  (27)    & 2159 (58317) & 2730 (73729)  & 286 (11.97\%) & 7730 (10.48\%) \\
SMAP  & 25  (54 )  & 2555 (138004)  & 8070 (435826)  & 1034 (12.40\%) & 55854 (12.82\%)\\
\hline
\end{tabular}
}
\caption{\small{The statistical profile of the datasets: MSL and SMAP.}
}
\label{tab:appendix_dataset_sts}
\end{table*}
Table~\ref{tab:simple_baseline_smap_msl_smd} summarizes the point-adjust F1 and standard F1 scores of simple baseline models and the published performance of the SOTA models. The performance of each proposed simple baseline model is averaged over all traces per dataset. The results of SOTA methods are taken from \citeauthor{kim2022TADevaluation}\citeyear{kim2022TADevaluation} in which only the best F1 scores are reported per method. The simple baselines, namely PCA-error and 1-NN distance, yield the best and second-best performance on both datasets, respectively.

\begin{table}[h]
\scalebox{0.78}{
\begin{tabular}{llllll} 
\hline
\multicolumn{2}{c}{\multirow{4}{*}{\textbf{Method}}} & \multicolumn{4}{c}{\textbf{Datasets}}\\
\multicolumn{2}{c}{} & \multicolumn{2}{c}{\textbf{MSL}} & \multicolumn{2}{c}{\textbf{SMAP}} \\
\multicolumn{2}{c}{} & $F1_{PA}$ & $F1$ & $F1_{PA}$ & $F1$ \\
\hline
\multirow{5}{*}{\rotatebox{90}{\parbox[inner-pos=t]{1.2cm}{\textbf{Simple baselines}}}} 
& Random & \textbf{0.931} & 0.190 & \textbf{0.961} & 0.227  \\
& Sensor range deviation & 0.441 & 0.328 & 0.389 & 0.273 \\
& L2-norm & 0.854 & 0.395 & 0.745 & 0.351  \\
& 1-NN distance & 0.912 &  \underline{0.404} & \underline{0.818} &  \underline{0.352}   \\
& PCA Error & 0.843 & \textbf{0.426} & 0.811 & \textbf{0.387} \\
\hline
\multirow{4}{*}{\rotatebox{90}{\parbox[inner-pos=t]{.4cm}{\textbf{SOTA Methods}}}} 
& DAGMM ~\cite{dagmm} & 0.701 & 0.199 & 0.712 & 0.333   \\
& OmniAnomaly ~\cite{OmniAnomaly} & 0.899 & 0.207 & 0.805 & 0.227   \\
& USAD ~\cite{usad} & \underline{0.927}  & 0.211 & \underline{0.818} & 0.228  \\
& GDN ~\cite{gdn} & 0.903 & 0.217 & 0.708 & 0.252  \\
\hline
\end{tabular} 
}
\caption{\small{Simple baselines outperform the SOTA deep-learning models on MSL and SMAP datasets. SOTA model performance is taken from \citeauthor{kim2022TADevaluation} (\citeyear{kim2022TADevaluation}). Bold: the best performance; underline: the second-best performance.}
}
\label{tab:simple_baseline_smap_msl_smd}
\end{table}

\begin{table*}[t]
\centering
\scalebox{0.62}{
\begin{tabular}{lrrrrrrrrrrrrrrrr}
\hline
\multicolumn{1}{c}{\multirow{2}{*}{\textbf{Method}}} & \multicolumn{16}{c}{\textbf{Datasets}}\\
{} & \multicolumn{4}{c}{ \textbf{SWAT}} & \multicolumn{4}{c}{\textbf{WADI 127}} & \multicolumn{4}{c}{\textbf{WADI 112}} & \multicolumn{4}{c}{\textbf{SMD}} \\
{} &     $F1$ &      $P$ &      $R$ &  $AUPRC$ &     $F1$ &      $P$ &      $R$ &  $AUPRC$ &   $F1$ &      $P$ &      $R$ &  $AUPRC$ &  $F1$ &      $P$ &      $R$ &  $AUPRC$ \\
\hline
MERLIN~\cite{Nakamura2020MERLINPD}             &  0.217 &  0.122 &  1.000 &  0.116 &  0.335 &  0.305 &  0.371 &  0.217 &    0.473 &  0.710 &  0.355 &  0.412 &  0.384 &  0.474 &  0.374 &  0.338 \\
DAGMM~\cite{dagmm}              &  0.770 &  0.991 &  0.630 &  0.727 &  0.279 &  0.993 &  0.162 &  0.207 &    0.520 &  0.932 &  0.361 &  0.469 &  0.435 &  0.564 &  0.497 &  0.370 \\
OmniAnomaly~\cite{omnianom}        &  0.773 &  0.990 &  0.634 &  0.736 &  0.281 &  1.000 &  0.163 &  0.212 &    0.441 &  0.607 &  0.346 &  0.441 &  0.415 &  0.566 &  0.464 &  0.360 \\
USAD~\cite{usad}               &  0.772 &  0.988 &  0.634 &  0.730 &  0.279 &  0.993 &  0.162 &  0.207 &    0.535 &  0.744 &  0.417 &  0.483 &  0.426 &  0.546 &  0.474 &  0.364 \\
GDN~\cite{gdn}                &  0.810 &  0.987 &  0.686 &  0.762 &  0.347 &  0.643 &  0.237 &  0.304 &    0.571 &  0.727 &  0.470 &  0.519 &  0.526 &  0.597 &  0.565 &  0.457 \\
TranAD~\cite{tuli2022tranad}             &  0.800 &  0.990 &  0.671 &  0.759 &  0.340 &  0.293 &  0.404 &  0.215 &    0.511 &  0.795 &  0.377 &  0.529 &  0.457 &  0.579 &  0.481 &  0.387 \\
AnomalyTransformer~\cite{xu2022anomaly} &  0.765 &  0.943 &  0.643 &  0.712 &  0.209 &  0.122 &  0.743 &  0.188 &    0.543 &  0.576 &  0.513 &  0.427 &  0.426 &  0.419 &  0.528 &  0.313 \\
\hline
Random                            &  0.218 &  0.122 &  0.997 &  0.121 &  0.101 &  0.053 &  0.958 &  0.054 &    0.101 &  0.053 &  0.992 &  0.053 &  0.080 &  0.044 &  0.696 &  0.043 \\
Sensor Range Deviation            &  0.231 &  0.131 &  0.979 &  0.556 &  0.101 &  0.053 &  1.000 &  0.317 &    0.465 &  0.567 &  0.394 &  0.497 &  0.132 &  0.110 &  0.682 &  0.321 \\
L2-norm                    &  0.782 &  0.985 &  0.648 &  0.715 &  0.281 &  1.000 &  0.163 &  0.210 &    0.513 &  0.887 &  0.361 &  0.474 &  0.404 &  0.569 &  0.455 &  0.343 \\
1-NN distance          &  0.782 &  0.984 &  0.649 &  0.726 &  0.281 &  1.000 &  0.163 &  0.211 &    0.568 &  0.779 &  0.447 &  0.501 &  0.463 &  0.626 &  0.458 &  0.389 \\
PCA Error      &  0.833 &  0.965 &  0.733 &  0.744 &  0.501 &  0.884 &  0.350 &  0.476  & 0.655 &  0.752 &  0.580 &  0.570 &  0.572 &  0.611 &  0.584 &  0.515 \\
\hline
1-Layer MLP &	0.771	& 0.981 &	0.635 &	0.797 &	0.267	& 0.834	& 0.159 &	0.193	& 0.502	& 0.880 & 	0.351	& 0.437	& 0.514 &	0.598	& 0.574 &	0.458 \\
Single block MLPMixer &	0.780 & 0.854 &	0.718 & 0.791 &	0.275 &	0.862 &	0.163 &	0.218 &	0.497 &	0.822 &	0.356 &	0.426 &	0.512 &	0.608 &	0.554 &	0.458 \\
Single Transformer block &	0.787 &	0.868 &	0.720 &	0.821 & 0.289 &	0.908 & 0.172 &	0.255 &	0.534 &	0.735 &	0.419 & 0.453 &	0.489 & 0.589 & 0.536 &	0.422 \\
1-Layer GCN-LSTM & 0.829 &	0.982 &	0.718 & 0.793 &	0.439 &	0.744 &	0.311 &	0.367 &	0.596 &	0.742 &	0.498 &	0.535 &	0.550 &	0.627 &	0.599 &	0.478 \\
\hline
\end{tabular} }
\caption{\small{Experimental results for SWaT, WADI, and SMD datasets evaluated under the standard point-wise metric.}}
\label{tab:sota-standard}
\end{table*}

\begin{table*}[t]
\centering
\scalebox{0.62}{
\begin{tabular}{lrrrrrrrrrrrrrrrr}
\hline
\multicolumn{1}{c}{\multirow{2}{*}{\textbf{Method}}} & \multicolumn{16}{c}{\textbf{Datasets}}\\
{} & \multicolumn{4}{c}{ \textbf{SWAT}} & \multicolumn{4}{c}{\textbf{WADI 127}} & \multicolumn{4}{c}{\textbf{WADI 112}} & \multicolumn{4}{c}{\textbf{SMD}} \\
{} &     $F1$ &      $P$ &      $R$ &  $AUPRC$ &    $F1$ &      $P$ &      $R$ &  $AUPRC$ & $F1$ &      $P$ &      $R$ &  $AUPRC$ & $F1$ &      $P$ &      $R$ &  $AUPRC$ \\
\hline
MERLIN~\cite{Nakamura2020MERLINPD}             &  0.286 &  0.521 &  0.197 &  0.180 &  0.354 &  0.308 &  0.416 &  0.247 &    0.503 &  0.748 &  0.379 &  0.466 &  0.473 &  0.641 &  0.407 &  0.406 \\
DAGMM~\cite{dagmm}              &  0.402 &  0.646 &  0.292 &  0.403 &  0.406 &  0.993 &  0.255 &  0.295 &    0.609 &  0.938 &  0.451 &  0.538 &  0.379 &  0.552 &  0.405 &  0.316 \\
OmniAnomaly~\cite{omnianom}        &  0.367 &  0.403 &  0.337 &  0.394 &  0.410 &  1.000 &  0.258 &  0.303 &    0.496 &  0.671 &  0.393 &  0.492 &  0.353 &  0.490 &  0.410 &  0.300 \\
USAD~\cite{usad}               &  0.413 &  0.674 &  0.298 &  0.408 &  0.406 &  0.993 &  0.255 &  0.295 &    0.573 &  0.754 &  0.462 &  0.524 &  0.364 &  0.539 &  0.375 &  0.303 \\
GDN~\cite{gdn}                &  0.385 &  0.418 &  0.357 &  0.423 &  0.434 &  0.799 &  0.298 &  0.348 &    0.588 &  0.812 &  0.461 &  0.543 &  0.570 &  0.673 &  0.550 &  0.500 \\
TranAD~\cite{tuli2022tranad}             &  0.425 &  0.388 &  0.471 &  0.464 &  0.353 &  0.301 &  0.425 &  0.239 &    0.589 &  0.795 &  0.468 &  0.604 &  0.390 &  0.544 &  0.399 &  0.322 \\
AnomalyTransformer~\cite{xu2022anomaly} &  0.331 &  0.885 &  0.204 &  0.348 &  0.219 &  0.128 &  0.738 &  0.189 &    0.555 &  0.589 &  0.524 &  0.451 &  0.351 &  0.350 &  0.460 &  0.247 \\ 
\hline
Random                            &  0.217 &  0.123 &  0.951 &  0.124 &  0.106 &  0.056 &  0.919 &  0.057 &    0.106 &  0.056 &  0.952 &  0.055 &  0.080 &  0.046 &  0.707 &  0.045 \\
Sensor Range Deviation            &  0.230 &  0.131 &  0.928 &  0.534 &  0.098 &  0.053 &  0.678 &  0.374 &    0.526 &  0.569 &  0.489 &  0.543 &  0.116 &  0.121 &  0.508 &  0.325 \\
L2-norm                    &  0.366 &  0.898 &  0.230 &  0.367 &  0.410 &  1.000 &  0.258 &  0.300 &    0.607 &  0.908 &  0.456 &  0.582 &  0.338 &  0.461 &  0.411 &  0.281 \\
1-NN distance         &  0.372 &  0.937 &  0.232 &  0.391 &  0.410 &  1.000 &  0.258 &  0.301 &    0.618 &  0.915 &  0.467 &  0.564 &  0.384 &  0.517 &  0.389 &  0.327 \\
PCA Error &  0.574 &  0.918 &  0.417 &  0.504 &  0.557 &  0.884 &  0.406 &  0.543 & 0.699 &  0.752 &  0.652 &  0.640 &  0.580 &  0.641 &  0.597 &  0.516\\

\hline
1-Layer MLP	& 0.519 & 0.740	& 0.400	& 0.532 & 0.384 &	0.834 &	0.250 &	0.280 &	0.558 &	0.880 &	0.408 &	0.493 &	0.487 &	0.536 &	0.513 &	0.424 \\
Single block MLPMixer &	0.549	& 0.762 &	0.430 &	0.549 &	0.396 &	0.862 &	0.257 &	0.307 &	0.552 &	0.865 &	0.405 &	0.484 &	0.472 &	0.525 &	0.556 &	0.426 \\
Single Transformer block &	0.526 &	0.556 &	0.500 &	0.573 &	0.416 &	0.908 &	0.270 &	0.354 &	0.575 &	0.735 &	0.472 &	0.506 &	0.420 &	0.531 &	0.439 &	0.370 \\
1-Layer GCN-LSTM &	0.532 &	0.914 &	0.375 &	0.532 &	0.540 &	0.745 &	0.424 &	0.468 &	0.645 &	0.742 &	0.570 &	0.599 &	0.535 &	0.591 &	0.566 &	0.462 \\
\hline
\end{tabular}}
\caption{\small{Experimental results for SWaT, WADI, and SMD datasets evaluated under the time-series range-wise metric.}}
\label{tab:sota-wagner}
\end{table*}


\begin{table*}[t]
\centering
\scalebox{0.62}{
\begin{tabular}{lrrrrrrrrrrrrrrrr}
\hline
\multicolumn{1}{c}{\multirow{2}{*}{\textbf{Method}}} & \multicolumn{16}{c}{\textbf{Datasets}}\\
{} & \multicolumn{4}{c}{ \textbf{UCR/IB-16}} & \multicolumn{4}{c}{\textbf{UCR/IB-17}} & \multicolumn{4}{c}{\textbf{UCR/IB-18}} & \multicolumn{4}{c}{\textbf{UCR/IB-19}} \\
{} &  $F1$ &  $P$ &  $R$ &  $AUPRC$ &     $F1$ &      $P$ &      $R$ &  $AUPRC$ &   $F1$ &      $P$ &      $R$ &  $AUPRC$ &  $F1$ &      $P$ &      $R$ &  $AUPRC$ \\
\hline
LOF~\cite{breunig2000lof} &0.476	&0.555	&0.416	&0.196	&0.959	&0.955	&0.963	&0.944	&0.916	&0.920	&0.911	&0.832	&0.857	&0.818	&0.900	&0.939 \\

MERLIN~\cite{Nakamura2020MERLINPD} & 0.846 & 	0.786 & 	0.917 & 	0.871 & 	0.987 & 	0.982 & 	0.991 & 	0.979 & 	0.795 & 	0.986 & 	0.667 & 	0.724 & 	0.870 & 	0.769 & 	1.000 & 	0.945 \\
\hline
Random & 	0.005	& 0.002	& 0.500	& 0.002	& 0.041	& 0.024	& 0.144	& 0.018	& 0.039	& 0.020	& 0.725	& 0.017	& 0.030	& 0.016	& 0.200	& 0.006 \\ 
Sensor range deviation &	0.004 &	0.002 &		1.000 &		0.001 &		0.085 &		0.200 &		0.054 &		0.136 &		0.038 &		0.020 &		1.000 &		0.010 &		0.004 &		0.002 &		1.000 &		0.001 \\
L2-norm	 & 0.011	 & 0.005	 & 1.000	 & 0.003	 & 0.058	 & 0.030	 & 0.748	 & 0.024	 & 0.061	 & 0.032	 & 0.794	 & 0.026	 & 0.017 & 	0.008 & 	1.000 & 	0.005 \\
1-NN distance  & 	0.786  & 	0.688  & 	0.917  & 	0.471  & 	0.973 & 	0.965	 & 0.982 & 	0.992 & 	0.889 & 	0.876 & 	0.902 & 	0.961 & 	0.870 & 	0.769 & 	1.000 & 	0.788 \\
PCA Error & 	0.750 & 	0.600 & 	1.000 & 	0.737 & 	0.974 & 	0.949 & 	1.000 & 	0.997	 & 0.990	 & 0.981	 & 1.000	 & 1.000	 & 1.000	 & 1.000	 & 1.000	 & 1.000 \\
\hline
\end{tabular} }
\caption{\small{Experimental results for four univariate UCR/InternalBleeding datasets evaluated under the standard point-wise metric.}}
\label{tab:ucr-standard}
\end{table*}


\begin{table*}[t]
\centering
\scalebox{0.62}{
\begin{tabular}{lrrrrrrrrrrrrrrrr}
\hline
\multicolumn{1}{c}{\multirow{2}{*}{\textbf{Method}}} & \multicolumn{16}{c}{\textbf{Datasets}}\\
{} & \multicolumn{4}{c}{ \textbf{UCR/IB-16}} & \multicolumn{4}{c}{\textbf{UCR/IB-17}} & \multicolumn{4}{c}{\textbf{UCR/IB-18}} & \multicolumn{4}{c}{\textbf{UCR/IB-19}} \\
{} &     $F1$ &      $P$ &      $R$ &  $AUPRC$ &     $F1$ &      $P$ &      $R$ &  $AUPRC$ &   $F1$ &      $P$ &      $R$ &  $AUPRC$ &  $F1$ &      $P$ &      $R$ &  $AUPRC$ \\
\hline
LOF~\cite{breunig2000lof} &0.476	&0.555	&0.416	&0.223	&0.955	&0.947	&0.964	&0.946	&0.911	&0.920	&0.902	&0.837	&0.857	&0.818	&0.900	&0.939 \\

MERLIN~\cite{Nakamura2020MERLINPD} & 	0.846 & 	0.786 & 	0.917 & 	0.872 & 	0.987 & 	0.982 & 	0.991 & 	0.981 & 	0.791 & 	0.986 & 	0.660 & 	0.763 & 	0.870 & 	0.769 & 	1.000 & 	0.941 \\
\hline
Random & 	0.030	 & 0.016	 & 0.229	 & 0.005	 & 0.116	 & 0.062	 & 0.947	 & 0.062 & 	0.091	 & 0.050	 & 0.469	 & 0.043	 & 0.048	 & 0.031	 & 0.100	 & 0.008 \\
Sensor range deviation &	0.000 &	0.000 &		1.000 &		0.001 &		0.094 &		0.353 &		0.054	 &	0.212	 &	0.000	 &	0.000	 &	1.000	 &	0.010	 &	0.000	 &	0.000	 &	1.000 &	0.001 \\
L2-norm	 &	0.021	 &	0.010	 &	1.000 &		0.006	 &	0.164	 &	0.092 &		0.734 &		0.076	 &	0.123 &		0.067 &		0.794 &		0.054 &		0.050	 &	0.026	 &	1.000	 &	0.016 \\
1-NN distance	 &	0.786	 &	0.688	 &	0.917	 &	0.480	 &	0.969	 &	0.957 &		0.982	 &	0.992	 &	0.902	 &	0.828	 &	0.990	 &	0.961	 &	0.870	 &	0.769	 &	1.000 &		0.791 \\
PCA Error	 &	0.750	 &	0.600	 &	1.000	 &	0.708	 &	0.974 &		0.949 &		1.000 &		0.997	 &	0.990	 &	0.981	 &	1.000 &		0.999 &		1.000	 &	1.000	 &	1.000	 &	1.000 \\
\hline
\end{tabular}}
\caption{\small{Experimental results for four univariate UCR/InternalBleeding datasets evaluated under the time-series range-wise metric.}}
\label{tab:ucr-wagner}
\end{table*}


\end{document}